\documentclass{article}
\usepackage{spconf,amsmath,graphicx}
\usepackage{amsfonts}
\usepackage{booktabs}
\usepackage[table]{xcolor}
\title{Simple Baselines for Projection-based Full-reference and No-reference Point Cloud Quality Assessment}
\name{Zicheng Zhang*$^1$, Yingjie Zhou*$^1$, Wei Sun$^1$, Xiongkuo Min$^1$, and Guangtao Zhai$^{1,2}$\thanks{*These authors contributed equally to this work. This work was supported in part by NSFC (No.62225112, No.61831015), the Fundamental Research Funds for the Central Universities, National Key R\&D Program of China 2021YFE0206700, Shanghai Municipal Science and Technology Major Project (2021SHZDZX0102), and STCSM 22DZ2229005. }}
\address{$^{1}$Institute of Image Communication and Network Engineering, Shanghai Jiao Tong University, China\\
$^{2}$MoE Key Lab of Artificial Intelligence, AI Institute, Shanghai Jiao Tong University, China}
\begin{document}
%\ninept
%
\maketitle
\begin{abstract}
Point clouds are widely used in 3D content representation and have various applications in multimedia. However, compression and simplification processes inevitably result in the loss of quality-aware information under storage and bandwidth constraints. Therefore, there is an increasing need for effective methods to quantify the degree of distortion in point clouds. In this paper, we propose simple baselines for projection-based point cloud quality assessment (PCQA) to tackle this challenge. We use multi-projections obtained via a common cube-like projection process from the point clouds for both full-reference (FR) and no-reference (NR) PCQA tasks. Quality-aware features are extracted with popular vision backbones. The FR quality representation is computed as the similarity between the feature maps of reference and distorted projections while the NR quality representation is obtained by simply squeezing the feature maps of distorted projections with average pooling The corresponding quality representations are regressed into visual quality scores by fully-connected layers. Taking part in the ICIP 2023 PCVQA Challenge, we succeeded in achieving the top spot in four out of the five competition tracks.
\end{abstract}
\begin{keywords}
Point cloud, quality assessment, projection-based, full-reference, no-reference
\end{keywords}
\section{Introduction}
\label{sec:intro}
Point clouds have emerged as an effective means of representing 3D content and have found extensive applications in immersive domains such as virtual reality \cite{xiong2021augmented}, mesh representation \cite{9722570}, and metaverse \cite{ning2021survey}. However, due to constraints in storage space and transmission bandwidth, point clouds are subject to lossy processes such as compression and simplification, which can result in the loss of quality-aware information to balance bit rates.  Consequently, there is a pressing need for methods that can effectively quantify the degree of distortion in point clouds, to enable the development of compression systems and enhance the Quality of Experience (QoE) for viewers.

% Moreover, distortions such as noise and blur may further compromise the visual quality of point clouds, stemming from sensor inaccuracies and rendering techniques.

According to the feature extraction types, the point cloud quality assessment (PCQA) methods can be categorized into model-based and projection-based methods. The model-based methods directly extract quality-aware features from the point clouds while the projection-based methods infer the visual quality of point clouds via the rendered projections. Additionally, the PCQA methods can also be divided into full-reference (FR), reduced-reference (RR), and no-reference (NR) methods according to the involved content of reference. Early FR-PCQA methods simply focus on the point level, which includes p2point \cite{p2point} and p2plane \cite{p2plane} However, these methods only take geometry information into consideration, thus 
some FR-PCQA methods such as PointSSIM \cite{alexiou2020pointssim}, GraphSIM \cite{yang2020graphsim}, and PCQM \cite{meynet2020pcqm} are proposed to predict the quality difference between the reference and distorted point clouds by including color features and taking advantage of various features. The NR-PCQA method 3D-NSS \cite{zhang2021nomesh,zhang2022no} uses several statistical distributions to estimate quality-aware parameters from the geometry and color attributes' distributions. Later, some researchers further propose to use 2D projections to evaluate the visual quality of point clouds, achieving competitive performance with the assistance of mature IQA methods. Namely, PQA-net \cite{liu2021pqa} involves extracting features through multi-view projection techniques. Meanwhile, Fan $et$ $al.$ \cite{fan2022no,zhang2022treating} evaluates the visual quality of point clouds by analyzing captured video sequences. More recently, MM-PCQA \cite{zhang2023mm} takes advantage of both point cloud and projections and extracts features from both modalities.

\begin{figure*}[!htp]
    \centering
    \includegraphics[width = \linewidth]{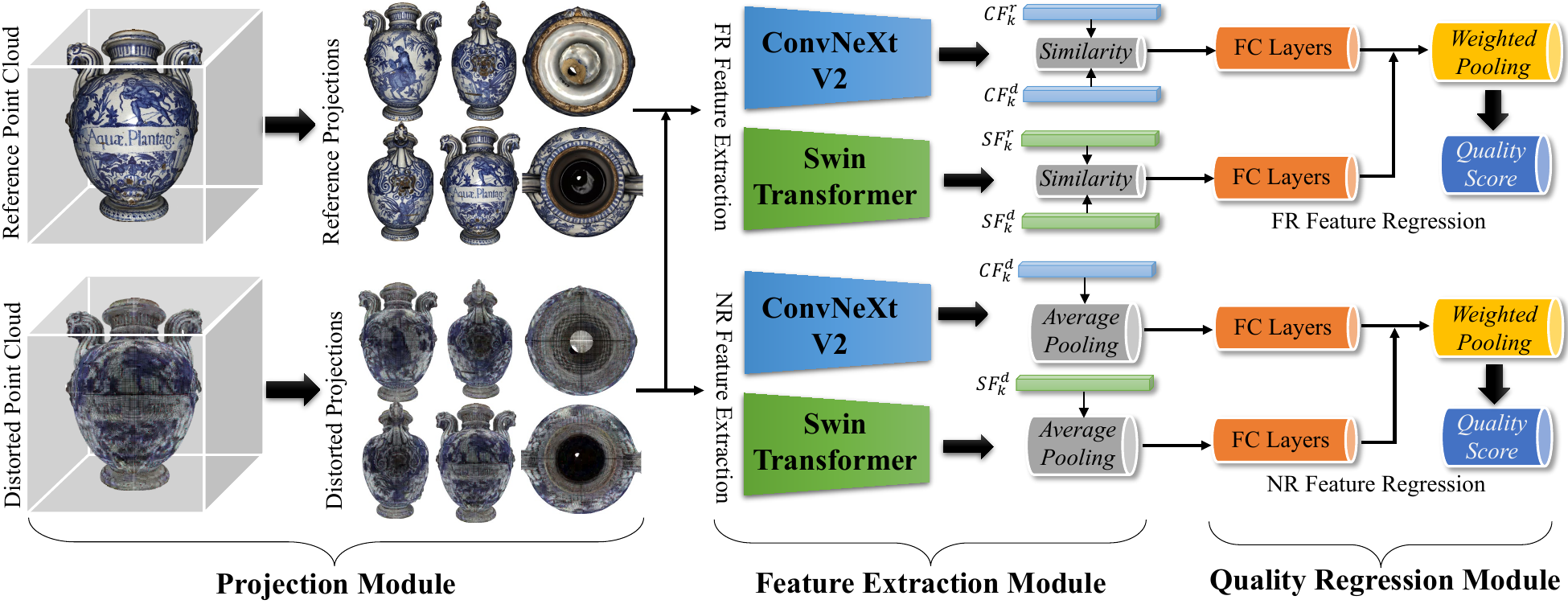}
    \caption{The framework of the proposed method. The projections are first rendered from the point clouds. Then the ConvNeXt V2 \cite{woo2023convnext} and the Swin Transformer \cite{liu2021swin} are employed to extract quality-aware information from the projections. Afterward, the extracted features are regressed into quality scores. In addition, the ConvNeXt V2 and the Swin Transformer backbones are trained separately. }
    \label{fig:framework}
\end{figure*}

Although the projection-based methods are highly dependent on the viewpoints, we can ease the viewpoint bias by employing multi-projections \cite{zhang2023gms,zhang2023advancing}. Furthermore, benefiting from the mature development of 2D vision backbones, the effectiveness and efficiency of the projection-based methods can be further boosted. Therefore, in this paper, we propose simple baselines for projection-based FR and NR PCQA.
Specifically, the common cube-like projection process is utilized to obtain multi-projections from the point clouds. Then the well-performing 2D vision backbones  ConvNeXt V2 \cite{woo2023convnext} and Swin Transformer \cite{liu2021swin} are both used to extract quality-aware features from the projections. For the FR baseline, the similarity between the feature maps of reference and distorted projections is computed as the quality representation. For the NR baseline, the feature maps of the distorted projections are simply squeezed into quality representation with average pooling. Finally, the FR and NR quality representations are regressed into visual quality scores with the assistance of fully-connected layers. 
Participating in the ICIP 2023 PCVQA Challenge, we emerged victorious in four out of the five competition tracks, which reveals that the proposed baselines are competitive for both FR-PCQA and NR-PCQA tasks.

\section{Proposed Method}
The framework of the proposed method is illustrated in Fig. \ref{fig:framework}, which includes the projection module, feature extraction module, and quality regression module. 

\subsection{Cube-Projection Process}
In order to ensure that we cover a wide range of viewing perspectives, we have chosen to use the widely-used cube-like viewpoints setting, which is also utilized in the popular MPEG VPCC point cloud compression standard \cite{graziosi2020overview}. Our approach involves using six different viewpoints that are perpendicular to each other, allowing us to capture rendered projections corresponding to the six surfaces of a cube, as illustrated in the projection module of Fig. \ref{fig:framework}.
Given a point cloud $\mathcal{P}$, the cube-projection process can be described as:

\begin{equation}
\begin{aligned}
     &\mathbf{P}  = \psi(\mathcal{P}), \\
    \mathbf{P}  = \{&{P}_{k}|k =1,\cdots, 6\},
\end{aligned}
\end{equation}
where $\mathbf{P}$ represents the set of the 6 rendered projections, $\psi(\cdot)$ stands for the rendering process, and ${P}_{k}$ indicates the $k$-th rendered projection. 

\subsection{Feature Extraction Module}
As stated in \cite{gu2022ntire}, the convolution network is better at retaining more spatial information while the transformer network can better capture global semantic features in the quality assessment tasks. Therefore, we propose to use the popular convolution network backbone ConvNeXt V2 \cite{woo2023convnext} and the transformer backbone Swin Transformer \cite{liu2021swin} to jointly extract quality-aware features from the projections.

% \in \mathbb{R}^{H_C \times W_C \times C_C} 

\subsubsection{FR Feature Extraction}
Given the projections sets $\mathbf{P}^r$ and $\mathbf{P}^d$ rendered from the reference and distorted point cloud pairs, we first extract the feature maps with the backbones introduced above:

\begin{equation}
\begin{aligned}
    &CF_{k}^r = \mathcal{C}({P}_{k}^r), SF_{k}^r = \mathcal{S}({P}_{k}^r),\\
    &CF_{k}^d = \mathcal{C}({P}_{k}^d), SF_{k}^d = \mathcal{S}({P}_{k}^d),
\end{aligned}
\end{equation}
where $\mathcal{C}(\cdot)$ and $\mathcal{S}(\cdot)$ represent the feature extraction operation of ConvNeXt V2 and Swin Transformer backbones, $CF_{k}^r $ and $SF_{k}^r$ stand for the extracted reference feature maps of the $k$-th reference projection ${P}_{k}^r$, and $CF_{k}^d $ and $SF_{k}^d$ stand for the extracted distorted feature maps of the $k$-th distorted projection ${P}_{k}^d$ respectively. Inspired by the perceptual similarity calculation form described in \cite{ding2020image}, we compute the structure and texture similarities as defined below:
\begin{equation}
\begin{aligned}
&\alpha (A_k, B_k)=\frac{2\mu_{A_k}\mu_{B_k} + \gamma_1}{(\mu_{A_k})^2+(\mu_{B_k})^2+\gamma_1}, \\
&\beta (A_k, B_k)=\frac{2\sigma_{A_k,B_k}+\gamma_2}{(\sigma_{A_k})^2+(\sigma_{B_k})^2+\gamma_2}, \\
& \overline{CF_k} = \alpha(CF_{k}^r,CF_{k}^d) \oplus \beta(CF_{k}^r,CF_{k}^d), \\
& \overline{SF_k} = \alpha(SF_{k}^r,SF_{k}^d) \oplus \beta(SF_{k}^r,SF_{k}^d), \\
\end{aligned}
\end{equation}
where $\alpha(\cdot)$ and $\beta(\cdot)$ indicate the texture and structure similarity calculation operation \cite{ding2020image}, $\mu_{F_{A_k}}, \mu_{F_{B_k}},(\sigma_{F_{A_k}})^2,(\sigma_{F_{A_k}})^2$, and $\sigma_{F_{A_k,B_k}}$ are the global means and variances of feature maps $A_k$ and $B_k$, and the global covariance between $A_k$ and $B_k$, $\oplus$ represents the concatenation operation, $\gamma_1$ and $\gamma_2$ are small constants to avoid instability, and $\overline{CF_k}$ and $\overline{SF_k}$ denote the final FR quality features extracted by the ConvNeXt V2 and Swin Transformer backbones from the $k$-th projections respectively. 
($CF_{k}^r$ and $CF_{k}^d$ $\in \mathbb{R}^{H_c \times W_c \times C_c}$, 
$SF_{k}^r$ and $SF_{k}^d$ $\in \mathbb{R}^{H_s \times W_s \times C_s}$, $\overline{CF_k} \in \mathbb{R}^{1 \times 2C_c}$, and $\overline{SF_k} \in \mathbb{R}^{1 \times 2C_s}$, where $C_c$ and $C_s$ stand for the number of channels for ConvNeXt V2 and Swin Transformer feature maps respectively.)

\subsubsection{NR Feature Extraction}
Given the projections set $\mathbf{P}^d$ rendered from the distorted point cloud, we can similarly obtain the quality-aware features:

\begin{equation}
\begin{aligned}
    &\widetilde{CF_k} = Avg(\mathcal{C}({P}_{k}^d)), \\
    &\widetilde{SF_k} = Avg(\mathcal{S}({P}_{k}^d)),
\end{aligned}
\end{equation}
where $\widetilde{CF_k}$ and $\widetilde{SF_k}$ represent the NR quality features extracted by the ConvNeXt V2 and Swin Transformer backbones from the $k$-th projection, and $Avg(\cdot)$ indicates the average pooling operation. ($\widetilde{CF_k} \in \mathbb{R}^{1 \times C_c} $ and $\widetilde{SF_k} \in \mathbb{R}^{1 \times C_s}$.)   

\subsection{Quality Regression Module}
Once the feature extraction module has extracted the quality-aware feature representation, we require a regression model to map these features to quality scores. To accomplish this, we utilize two-layer, fully connected (FC) layers to obtain the projection-level quality scores, which are consequently averaged into the final point cloud quality scores:
\begin{equation}
\begin{aligned}
    &Q_{FR} = \omega_{c} \frac{1}{6} \sum_{k=1}^6 \mathcal{FC}(\overline{CF_k}) + \omega_{s} \frac{1}{6} \sum_{k=1}^6 \mathcal{FC}(\overline{SF_k}), \\
    &Q_{NR} = \omega_{c} \frac{1}{6} \sum_{k=1}^6 \mathcal{FC}(\widetilde{CF_k}) + \omega_{s} \frac{1}{6} \sum_{k=1}^6 \mathcal{FC}(\widetilde{SF_k}),
\end{aligned}
\end{equation}
where $\mathcal{FC}(\cdot)$ stands for the FC layers regression operation, $\omega_{c}$ and $\omega_{s}$ are the weights to control the contribution proportion of the ConvNeXt V2 and Swin Transformer, and $Q_{FR}$ and $Q_{NR}$ are the corresponding predicted FR and NR quality scores. The Mean Squared Error (MSE) is utilized as the loss function:
\begin{equation}
    Loss = \frac{1}{n}\sum_{m=1}^{n}\left(Q-Q_{label}\right)^{2},
\end{equation}
where $n$ indicates the number of point clouds in a mini-batch, $Q$ and $Q_{label}$ are the predicted quality levels and subjective quality labels respectively.

\section{Experiment}
\subsection{Database \& Evaluation Criteria}
We participate in all 5 tracks of the ICIP 2023 PCVQA Challenge. 
The proposed method is validated on the BASICS database \cite{ak2023basics}, which is targeted at the quality assessment of compressed point clouds. The BASICS database contains 75 reference point clouds and employs 4 types of compression methods to generate the compressed point clouds. 

5 criteria are included to evaluate the performance, which consists of Pearson Linear Correlation Coefficient (PLCC), Spearman Rank Order Correlation Coefficient (SRCC), Difference/Similar Analysis quantified by Area Under the Curve (D/S$_{auc}$) \cite{krasula2016accuracy}, Better/Worse Analysis quantified by Correct Classification percentage (B/W$_{cc}$) \cite{krasula2016accuracy}, and Runtime Complexity (RC).

\subsection{Implementation Details}
The ConvNeXt V2 \cite{woo2023convnext} base and the Swin Transformer \cite{liu2021swin} base are selected as the backbones, which are both initialized with the weights pretrained on the ImageNet-22K database \cite{russakovsky2015imagenet}. The white background of the projections is removed. Then we resize the resolution of the minimum dimension of the projections
as 520 while maintaining their aspect ratios and the 384$\times$384 patches are cropped as the input. The Adam optimizer \cite{kingma2014adam} with the initial learning rate 4e-5 is utilized and the batch size is set as 6. The two backbones are trained separately and the proposed method is trained on a server with NVIDIA 3090.

Specifically, the BASICS database provides a fixed train-validation-test split. We conduct a $k$-fold training strategy on the training sets and evaluate the performance on the validation set. The top-performing models are then saved for evaluation on the testing set. The testing set is not available during the development phase. The final competition results are only based on the performance on the testing set.

\begin{table}[t]\small
\renewcommand\tabcolsep{1.5pt}
\setlength{\abovecaptionskip}{-5pt}
  \caption{Competition results on Track 1 (FR broad range) of the ICIP 2023 PCVQA Challenge. We are the team SJTU MMLAB and we achieve the 3rd place. Our performance is marked in gray.}
  \vspace{-0.2cm}
  \begin{center}
  \begin{tabular}{c|c|ccccc}
    \toprule
     Rank & Team &PLCC$\uparrow$  & SRCC$\uparrow$ & D/S$_{auc}$ $\uparrow$ & B/W$_{cc}$ $\uparrow$ & RC $\downarrow$\\ \hline
    1 & KDDIUSCJoint & 0.8754 & 0.9171	 & 0.8884 & 0.9696	& 42.80\\
    2 & CWI$\_$DIS & 0.8736 & 0.9090 & 0.8709 & 0.9607 & 1000.00\\
    \rowcolor{gray!40} 3 &  SJTU MMLAB  &0.8706 &0.8955 &0.8317 &0.9549 & 8.60 \\
    4 & SlowHand & 0.7911 & 0.8252 & 0.8045 & 0.9235 & 130.47\\
    \bottomrule
  \end{tabular}
  \end{center}
  \label{tab:tr1}
  \vspace{-0.7cm}
\end{table}

\begin{table}[!ht]\small
\renewcommand\tabcolsep{1.5pt}
\setlength{\abovecaptionskip}{-5pt}
  \caption{Competition results on Track 2 (NR broad range) of the ICIP 2023 PCVQA Challenge. Our team is the winner. }
  \vspace{-0.2cm}
  \begin{center}
  \begin{tabular}{c|c|ccccc}
    \toprule
     Rank & Team &PLCC$\uparrow$  & SRCC$\uparrow$ & D/S$_{auc}$ $\uparrow$ & B/W$_{cc}$ $\uparrow$ & RC $\downarrow$\\ \hline
    \rowcolor{gray!40} 1 &  SJTU MMLAB  &0.8806 &0.9076 &0.8481 &0.9626 & 16.10 \\
    2 & Q\&A & 0.7933 & 0.8038	 & 0.7878 & 0.9079	& 27.70\\
    3 & KDDIUSCJoint & 0.7595 & 0.7950	 & 0.7317 & 0.8911	& 11.53\\
    4 & Ecole des Mines & 0.5473 & 0.5883 & 0.6554 & 0.7764 & 5.53\\
    5 & SlowHand & 0.7911 & 0.8252 & 0.8045 & 0.9235 & 16.37\\
    \bottomrule
  \end{tabular}
  \end{center}
  \label{tab:tr2}
  \vspace{-0.7cm}
\end{table}

\begin{table}[!ht]\small
\renewcommand\tabcolsep{1.5pt}
\setlength{\abovecaptionskip}{-5pt}
  \caption{Competition results on Track 3 (FR high range) of the ICIP 2023 PCVQA Challenge. Our team is the winner. }
  \vspace{-0.2cm}
  \begin{center}
  \begin{tabular}{c|c|ccccc}
    \toprule
     Rank & Team &PLCC$\uparrow$  & SRCC$\uparrow$ & D/S$_{auc}$ $\uparrow$ & B/W$_{cc}$ $\uparrow$ & RC $\downarrow$\\ \hline
    \rowcolor{gray!40} 1 &  SJTU MMLAB  &0.6296 &0.5917 &0.6654 &0.9090 & 8.60 \\
    2 & KDDIUSCJoint & 0.5505 & 0.5160	 & 0.6420 & 0.8721	& 42.80\\
    3 & CWI$\_$DIS & 0.6029 & 0.4788 & 0.6250 & 0.8855 & 1000.00\\ 
    4 & SlowHand & 0.3768 & 0.4226 & 0.5654 & 0.7801 & 130.47\\
    \bottomrule
  \end{tabular}
  \end{center}
  \label{tab:tr3}
  \vspace{-0.7cm}
\end{table}

\begin{table}[!ht]\small
\renewcommand\tabcolsep{1.5pt}
\setlength{\abovecaptionskip}{-5pt}
  \caption{Competition results on Track 4 (NR high range) of the ICIP 2023 PCVQA Challenge. Our team is the winner. }
  \vspace{-0.2cm}
  \begin{center}
  \begin{tabular}{c|c|ccccc}
    \toprule
     Rank & Team &PLCC$\uparrow$  & SRCC$\uparrow$ & D/S$_{auc}$ $\uparrow$ & B/W$_{cc}$ $\uparrow$ & RC $\downarrow$\\ \hline
    \rowcolor{gray!40} 1 &  SJTU MMLAB  &0.6352 &0.6103 &0.6782 &0.9141 & 16.10 \\
    2 & Q\&A & 0.5526 & 0.4064	 & 0.6250 & 0.8691	& 27.70\\
    3 & KDDIUSCJoint & 0.4440 & 0.4167	 & 0.5790 & 0.7959	& 11.53\\
    4 & Ecole des Mines & 0.2761 & 0.1458 & 0.4939 & 0.6744 & 5.53\\
    5 & SlowHand & 0.0958 & 0.1065 & 0.4951 & 0.5569 & 16.37\\
    \bottomrule
  \end{tabular}
  \end{center}
  \label{tab:tr4}
  \vspace{-0.7cm}
\end{table}

\begin{table}[!ht]\small
\renewcommand\tabcolsep{1.5pt}
\setlength{\abovecaptionskip}{-5pt}
  \caption{Competition results on Track 5 (FR intra-reference) of the ICIP 2023 PCVQA Challenge. Our team is the winner. }
  \vspace{-0.2cm}
  \begin{center}
  \begin{tabular}{c|c|ccc}
    \toprule
     Rank & Team  & D/S$_{auc}$ $\uparrow$ & B/W$_{cc}$ $\uparrow$ & RC $\downarrow$\\ \hline
    \rowcolor{gray!40} 1 &  SJTU MMLAB  &0.8079 &0.9471 & 8.60 \\
    2 & KDDIUSCJoint  &0.8216 & 0.9330	& 42.80\\
    3 & CWI$\_$DIS & 0.8106 & 0.9384 & 1000.00\\ 
    4 & SlowHand &0.7533 & 0.8542 & 130.47\\
    \bottomrule
  \end{tabular}
  \end{center}
  \label{tab:tr5}
  \vspace{-0.7cm}
\end{table}

\subsection{Experiment Performance}
The competition results for all five tracks are exhibited in Table \ref{tab:tr1}, Table \ref{tab:tr2}, Table \ref{tab:tr3}, Table \ref{tab:tr4}, and Table \ref{tab:tr5} respectively, from which we can make several observations. 
a) The proposed method achieves 1st place in Tracks 2-4 in terms of PLCC and gains the 3rd place in the Track 1; b) The proposed method outperforms all the compared teams by a large performance margin in the NR tracks, which shows the superiority of the proposed method for the NR-PCQA tasks; c) The proposed method consumes much less time than the compared teams on the FR tracks, which is even about 7x times faster than the second RC ranking competitor (8.60 vs. 42.80). In all, the proposed method is both effective and efficient for both FR-PCQA and NR-PCQA tasks. Additionally, with the development of 2D vision backbones, the effectiveness can be further boosted. The proposed framework can also adopt lightweight vision backbones to adapt to the application scenario where computation resources are limited.

\section{Conclusion}
In conclusion, this paper proposes simple yet effective baselines for point cloud quality assessment (PCQA) through cube-like projection and feature extraction using popular vision backbones. Our approach utilizes multi-projections to generate full-reference (FR) and no-reference (NR) quality representations and regresses the quality representations into visual quality scores through fully-connected layers. The experimental results demonstrate the competitive performance of our proposed baselines for both FR and NR PCQA tasks in the ICIP 2023 PCVQA Challenge. Our work paves the way for future research to enhance the compression and simplification processes while improving the Quality of Experience (QoE) of viewers for point clouds.
\bibliographystyle{IEEEbib}
\bibliography{strings,refs}

\begin{thebibliography}{10}

\bibitem{xiong2021augmented}
Jianghao Xiong, En-Lin Hsiang, Ziqian He, Tao Zhan, and Shin-Tson Wu,
\newblock ``Augmented reality and virtual reality displays: emerging
  technologies and future perspectives,''
\newblock {\em Light: Science \& Applications}, vol. 10, no. 1, pp. 1--30,
  2021.

\bibitem{9722570}
Zhongpai Gao, Junchi Yan, Guangtao Zhai, Juyong Zhang, and Xiaokang Yang,
\newblock ``Robust mesh representation learning via efficient local
  structure-aware anisotropic convolution,''
\newblock {\em IEEE TNNLS}, pp. 1--13, 2022.

\bibitem{ning2021survey}
Huansheng Ning, Hang Wang, Yujia Lin, Wenxi Wang, Sahraoui Dhelim, Fadi Farha,
  Jianguo Ding, and Mahmoud Daneshmand,
\newblock ``A survey on metaverse: the state-of-the-art, technologies,
  applications, and challenges,''
\newblock {\em arXiv preprint arXiv:2111.09673}, 2021.

\bibitem{p2point}
P.~Cignoni, C.~Rocchini, and R.~Scopigno,
\newblock ``Metro: Measuring error on simplified surfaces,''
\newblock {\em Computer Graphics Forum}, vol. 17, no. 2, pp. 167--174, 1998.

\bibitem{p2plane}
Rufael Mekuria and Pablo Cesar,
\newblock ``Mp3dg-pcc, open source software framework for implementation and
  evaluation of point cloud compression,''
\newblock 2016, p. 1222–1226, Association for Computing Machinery.

\bibitem{alexiou2020pointssim}
Evangelos Alexiou and Touradj Ebrahimi,
\newblock ``Towards a point cloud structural similarity metric,''
\newblock in {\em IEEE ICMEW}. IEEE, 2020, pp. 1--6.

\bibitem{yang2020graphsim}
Qi~Yang, Zhan Ma, Yiling Xu, Zhu Li, and Jun Sun,
\newblock ``Inferring point cloud quality via graph similarity,''
\newblock {\em IEEE Transactions on Pattern Analysis and Machine Intelligence},
  2020.

\bibitem{meynet2020pcqm}
Gabriel Meynet, Yana Nehm{\'e}, Julie Digne, and Guillaume Lavou{\'e},
\newblock ``Pcqm: A full-reference quality metric for colored 3d point
  clouds,''
\newblock in {\em IEEE QoMEX}. IEEE, 2020, pp. 1--6.

\bibitem{zhang2021nomesh}
Zicheng Zhang, Wei Sun, Xiongkuo Min, Tao Wang, Wei Lu, Wenhan Zhu, and
  Guangtao Zhai,
\newblock ``A no-reference visual quality metric for 3d color meshes,''
\newblock in {\em ICMEW}. IEEE, 2021, pp. 1--6.

\bibitem{zhang2022no}
Zicheng Zhang, Wei Sun, Xiongkuo Min, Tao Wang, Wei Lu, and Guangtao Zhai,
\newblock ``No-reference quality assessment for 3d colored point cloud and mesh
  models,''
\newblock {\em IEEE TCSVT}, 2022.

\bibitem{liu2021pqa}
Qi~Liu, Hui Yuan, Honglei Su, Hao Liu, Yu~Wang, Huan Yang, and Junhui Hou,
\newblock ``Pqa-net: Deep no reference point cloud quality assessment via
  multi-view projection,''
\newblock {\em IEEE TCSVT}, vol. 31, no. 12, pp. 4645--4660, 2021.

\bibitem{fan2022no}
Yu~Fan, Zicheng Zhang, Wei Sun, Xiongkuo Min, Ning Liu, Quan Zhou, Jun He,
  Qiyuan Wang, and Guangtao Zhai,
\newblock ``A no-reference quality assessment metric for point cloud based on
  captured video sequences,''
\newblock in {\em IEEE MMSP}. IEEE, 2022, pp. 1--5.

\bibitem{zhang2022treating}
Zicheng Zhang, Wei Sun, Yucheng Zhu, Xiongkuo Min, Wei Wu, Ying Chen, and
  Guangtao Zhai,
\newblock ``Treating point cloud as moving camera videos: A no-reference
  quality assessment metric,''
\newblock {\em arXiv preprint arXiv:2208.14085}, 2022.

\bibitem{woo2023convnext}
Sanghyun Woo, Shoubhik Debnath, Ronghang Hu, Xinlei Chen, Zhuang Liu, In~So
  Kweon, and Saining Xie,
\newblock ``Convnext v2: Co-designing and scaling convnets with masked
  autoencoders,''
\newblock {\em arXiv preprint arXiv:2301.00808}, 2023.

\bibitem{liu2021swin}
Ze~Liu, Yutong Lin, Yue Cao, Han Hu, Yixuan Wei, Zheng Zhang, Stephen Lin, and
  Baining Guo,
\newblock ``Swin transformer: Hierarchical vision transformer using shifted
  windows,''
\newblock in {\em IEEE/CVF CVPR}, 2021, pp. 10012--10022.

\bibitem{zhang2023gms}
Zicheng Zhang, Wei Sun, Houning Wu, Yingjie Zhou, Chunyi Li, Xiongkuo Min,
  Guangtao Zhai, and Weisi Lin,
\newblock ``Gms-3dqa: Projection-based grid mini-patch sampling for 3d model
  quality assessment,''
\newblock {\em arXiv preprint arXiv:2306.05658}, 2023.

\bibitem{zhang2023advancing}
Zicheng Zhang, Wei Sun, Yingjie Zhou, Haoning Wu, Chunyi Li, Xiongkuo Min,
  Xiaohong Liu, Guangtao Zhai, and Weisi Lin,
\newblock ``Advancing zero-shot digital human quality assessment through
  text-prompted evaluation,''
\newblock {\em arXiv preprint arXiv:2307.02808}, 2023.

\bibitem{graziosi2020overview}
D~Graziosi, O~Nakagami, S~Kuma, A~Zaghetto, T~Suzuki, and A~Tabatabai,
\newblock ``An overview of ongoing point cloud compression standardization
  activities: Video-based (v-pcc) and geometry-based (g-pcc),''
\newblock {\em APSIPA Transactions on Signal and Information Processing}, vol.
  9, 2020.

\bibitem{gu2022ntire}
Jinjin Gu, Haoming Cai, Chao Dong, Jimmy~S Ren, Radu Timofte, Yuan Gong,
  Shanshan Lao, Shuwei Shi, Jiahao Wang, Sidi Yang, et~al.,
\newblock ``Ntire 2022 challenge on perceptual image quality assessment,''
\newblock in {\em IEEE/CVF CVPR}, 2022, pp. 951--967.

\bibitem{ding2020image}
Keyan Ding, Kede Ma, Shiqi Wang, and Eero~P Simoncelli,
\newblock ``Image quality assessment: Unifying structure and texture
  similarity,''
\newblock {\em IEEE TPAMI}, vol. 44, no. 5, pp. 2567--2581, 2020.

\bibitem{ak2023basics}
Ali Ak, Emin Zerman, Maurice Quach, Aladine Chetouani, Aljosa Smolic, Giuseppe
  Valenzise, and Patrick~Le Callet,
\newblock ``Basics: Broad quality assessment of static point clouds in
  compression scenarios,''
\newblock {\em arXiv preprint arXiv:2302.04796}, 2023.

\bibitem{krasula2016accuracy}
Luk{\'a}{\v{s}} Krasula, Karel Fliegel, Patrick Le~Callet, and Milo{\v{s}}
  Kl{\'\i}ma,
\newblock ``On the accuracy of objective image and video quality models: New
  methodology for performance evaluation,''
\newblock in {\em IEEE QoMEX}. IEEE, 2016, pp. 1--6.

\bibitem{russakovsky2015imagenet}
Olga Russakovsky, Jia Deng, Hao Su, Jonathan Krause, Sanjeev Satheesh, Sean Ma,
  Zhiheng Huang, Andrej Karpathy, Aditya Khosla, Michael Bernstein, et~al.,
\newblock ``Imagenet large scale visual recognition challenge,''
\newblock {\em IJCV}, vol. 115, no. 3, pp. 211--252, 2015.

\bibitem{kingma2014adam}
Diederik~P Kingma and Jimmy Ba,
\newblock ``Adam: A method for stochastic optimization,''
\newblock {\em ICLR}, 2014.

\end{thebibliography}

\end{document}